\documentclass[runningheads]{llncs}

\usepackage[hyphens]{url}  
\usepackage{graphicx} 
\usepackage[utf8]{inputenc} 
\usepackage{hyperref}       
\usepackage{url}            

\usepackage{booktabs}       
\usepackage{amsfonts}       
\usepackage{microtype}      
\usepackage{multirow}
\usepackage{graphicx}
\usepackage{amsmath,amsfonts,amssymb}
\usepackage{algorithm}
\usepackage[noend]{algpseudocode}

\usepackage{ulem}
\usepackage{subfig}
\usepackage{amssymb}
\usepackage{amsfonts}

\newcommand\numberthis{\addtocounter{equation}{1}\tag{\theequation}}

\title{Unsupervised Training for Neural TSP Solver\thanks{This research is funded by the Latvian Council of Science, project lzp-2021/1-0479.}}

\author{Elīza Gaile\inst{1,}\thanks{Corresponding author.} \and
Andis Draguns\inst{2} \and
Emīls Ozoliņš\inst{3} \and
Kārlis Freivalds\inst{4} }

\institute{Faculty of Computing, University of Latvia, Riga, Latvia \\ \email{eliiza.gaile@gmail.com} \and
Institute of Mathematics and Computer Science at University of Latvia, Riga, Latvia
\email{andis.draguns@lumii.lv} \and
\email{ozolinsemils@gmail.com} \and
Institute of Electronics and Computer Science, Riga, Latvia\\
\email{karlis.freivalds@edi.lv}}

\begin{document}

\maketitle

\begin{abstract}

There has been a growing number of machine learning methods for approximately solving the travelling salesman problem. However, these methods often require solved instances for training or use complex reinforcement learning approaches that need a large amount of tuning. To avoid these problems, we introduce a novel unsupervised learning approach. We use a relaxation of an integer linear program for TSP to construct a loss function that does not require correct instance labels. With variable discretization, its minimum coincides with the optimal or near-optimal solution. Furthermore, this loss function is differentiable and thus can be used to train neural networks directly. We use our loss function with a Graph Neural Network and design controlled experiments on both Euclidean and asymmetric TSP. Our approach has the advantage over supervised learning of not requiring large labelled datasets. In addition, the performance of our approach surpasses reinforcement learning for asymmetric TSP and is comparable to reinforcement learning for Euclidean instances. Our approach is also more stable and easier to train than reinforcement learning.

\end{abstract}

\section{Introduction}
Traveling salesman problem (TSP) is a well-known combinatorial optimization problem that searches for the optimal way to traverse a graph while visiting each node exactly once. TSP and its variants have broad practical applications, e.g. in electronics and logistics \cite{Matai2010TravelingSP}. Considering that TSP is an NP-hard problem, many analytical methods and heuristics have been handcrafted to solve this problem as optimally and efficiently as possible.

Neural networks have become powerful tools to solve various tasks and have shown encouraging results also for TSP. Training neural networks to solve combinatorial optimization tasks such as TSP presents distinct challenges for all learning paradigms – supervised (SL), unsupervised (UL), and reinforcement learning (RL). Recently, both supervised and reinforcement learning has been widely used to solve TSP, however, both of them have disadvantages.

While SL can perform better than other learning paradigms on fixed-size graphs (e.g. \cite{joshi2019efficient}), supervised learning requires optimally labeled TSP examples that are time consuming to produce even for moderately sized instances; besides that supervised learning cannot process multiple correct solutions. Although reinforcement learning does not have these problems, RL systems are often complex, unstable, and less sample efficient than other learning paradigms. Besides that, RL models are not well-suited for non-autoregressive models since that would require a vast action space with $O(n^2)$ continuous values. But autoregressive RL models rely on decoders that use node embeddings instead of adjacency matrix embeddings. This means that to solve asymmetric TSP using RL, the full adjacency matrix representation must be encoded in node embeddings. It can be done but requires powerful and complex encoders \cite{Kwon2021MatrixEN}.

We propose to train Neural models for TSP in an unsupervised way through minimization of a differentiable loss. By exchanging supervised loss function with this unsupervised loss, it is possible to eliminate the need for labeled training data without using RL and hence avoid RL disadvantages as well. The loss function is constructed so that its global minimum corresponds to the optimal solution of relaxed TSP; we use variable discretization to obtain integer solutions. Since loss function is non-convex, direct minimization for a given TSP instance usually ends up a sub-optimal local minimum, but when used for training a neural network, the trained network manages to find close-to-optimal solutions. 

The proposed loss function does not rely on labels so it can be applied on large instances while relieving us from the need to create a pre-solved dataset and does not have problems with multiple solutions. In addition, our unsupervised approach performs similarly to reinforcement learning when solving Euclidean instances and our approach works on asymmetric TSP which reinforcement learning struggles with.

\section{Related Work}
Over the last few years, many new machine learning approaches for solving TSP have been proposed. These approaches can be divided into two directions – hybrid methods where machine learning assists established classical heuristics and end-to-end neural solvers where ML model outputs solutions directly from the input.

Most of the notable end-to-end advances in TSP are based on sequence-to-sequence models \cite{Bello2017NeuralCO,Vinyals2015PointerN}, attention models \cite{Deudon2018LearningHF,Kool2019AttentionLT,Nazari2018ReinforcementLF} or graph neural networks \cite{joshi2019efficient,Khalil2017LearningCO,Kwon2021MatrixEN}. Earlier works focused more on  supervised learning (mostly based on Pointer Networks \cite{Vinyals2015PointerN}), however after that more and more RL approaches emerged (e.g. \cite{Bello2017NeuralCO,joshi2019efficient,Kool2019AttentionLT,Kwon2021MatrixEN}) and it was noted that reinforcement learning is generally more suitable for TSP than SL \cite{joshi2020learning,Joshi2019OnLP}. 

To our knowledge, there are no notable works on neural end-to-end TSP solvers with unsupervised learning despite UL also having the advantage of not requiring labeled data. There is also very little work on non-Euclidean TSP variants; most of the models used for solving TSP use only node coordinates without adjacency matrices as input and are only applied to planar TSP variants (e.g. \cite{Deudon2018LearningHF,joshi2019efficient,Kool2019AttentionLT,Vinyals2015PointerN}).

\section{Unsupervised TSP}
To train a neural network in an unsupervised manner, an unsupervised loss function is needed. We construct a differentiable function $\mathcal{L}: ([0,1]^{n \times n}, \mathbb{R}^{n \times n}) \rightarrow \mathbb{R}$, where $n$ – number of nodes in a TSP instance. The input of this function is two matrices – matrix $X$ which tells if each edge is a part of the proposed optimal tour ($0$ – not a part of the tour, $1$ – part of the tour) and matrix $C$ which is the adjacency matrix of the instance and contains weights of each edge. The output of $\mathcal{L}$ is an abstract numerical evaluation of how close the proposed tour is to the optimal solution; the smaller the output, the better.

\subsection{Unsupervised Loss}
Our approach to obtain such a loss function is based on a relaxation of an integer program for TSP that was first used by Dantzig, Fulkerson, and Johnson \cite{2003ImplementingDFJ}. If the vertices are numbered from $1$ to $n$, the TSP problem can be formulated as follows:
\begingroup
\allowdisplaybreaks
\begin{align*} \numberthis
\min \sum_{i=1}^n \sum_{j\ne i,j=1}^n c_{ij} \cdot x_{ij}&\colon &&  \\
      0  \leq x_{ij} &\leq 1 &&\\
      \sum_{i=1,i\ne j}^n x_{ij} &= 1 && j=1, \ldots, n; \\
      \sum_{j=1,j\ne i}^n x_{ij} &= 1 && i=1, \ldots, n; \\
     \sum_{i \in Q}{\sum_{j \ne i, j \in Q}{x_{ij}}} &\leq |Q| - 1 && \forall Q \subsetneq \{1, \ldots, n\}, |Q| \geq 2 \\
\end{align*}
\endgroup

The last constraint guarantees that the solution has no tours smaller than the whole graph and therefore is called the subtour constraint. The subtour constraint checks $O(2^n)$ node subsets. To avoid this exponential growth, the subtour constraint can be replaced by a heuristic algorithm that looks only at a part of subsets (the ones which are more likely to violate the constraint) and can therefore run in polynomial time. If the chosen heuristic is re-applied many times and each time the found constraint violations are corrected, a solution with no violations will be achieved.

For the heuristic, the parametric connectivity \cite{2003ImplementingDFJ} is used. First, this heuristic replaces the previous subtour contraint with
\begin{align*} \numberthis
\mathrm{cut}(Q, V-Q)  \geq 1 && \forall Q \subset \{1, \ldots, n\}, |Q| \geq 2,
\end{align*}
where $V$ - the set of all vertices and $\mathrm{cut}(S, T) = \sum_{i \in S}{\sum_{j \in T}{x_{ij}}}$. Similarly to DFJ, this constraint also ensures that there are no subtours in the TSP solution.

Second, the heuristic imagines the proposed solution $X$ as a separate graph with edge weights corresponding to $x_{ij}$ values (these are values that determine whether edge is is the solution). The heuristic calculates and then orders this graph's subsets by a parameter $\epsilon$, which is the maximum edge weight such that for any two vertices in the subset, they are connected by a path in with all edges of weight at least $\epsilon$. Assuming that the first three constraints of DFJ are satisfied, subsets of the graph with the largest $\epsilon$ have the smallest cuts (and the largest expected values to violate subtour constraint), therefore $n$ subsets with the largest $\epsilon$ parameters are chosen to be checked.

By combining the relaxation of the integer program and the parametric connectivity heuristic, the loss function for TSP with solution matrix $X$ and adjacency matrix $C$ is obtained. We evaluate the length of the tour and how much each of the DFJ constraints are violated ($0$ corresponds to satisfied constraints):
\begingroup
\begin{align*} \numberthis
\mathcal{L} = & \: \alpha \cdot  \sum_{i=1}^n\sum_{j=1, j\ne i}^n c_{ij} \cdot x_{ij} + \\
+ & \: \beta \cdot \left[ \sum_{j=1}^n {\left(1-\sum_{i=1, i\ne j}^n x_{ij}\right)}^2+\sum_{i=1}^n {\left(1-\sum_{j=1, j\ne i}^n x_{ij}\right)}^2 \right] + \\
+ & \: \gamma \cdot \sum_{Q \in S} \left(1-\mathrm{cut}(Q, V-Q)\right)^2
\end{align*}
\endgroup

where $S$ - node subsets chosen by the parametric connectivity heuristic that violate the subtour constraint; $\alpha$, $\beta$, $\gamma$ – scalars for scaling.

When the summands of the loss function are scaled appropriately, the global minimum of the function thus created satisfies all constraints and has the shortest tour, and can therefore express the quality of a proposed TSP solution without optimal labels. For this paper, we experimentally found that values $\alpha = 5$, $\beta =1$, $\gamma = 1$ work well.

Additionally, the proposed loss function $\mathcal{L}$ is differentiable w. r. t. solution $X$ and can therefore be used to train a neural network directly. It is not differentiable w. r. t. to the choice of node subsets $S$ and if we wanted to perfectly evaluate the proposed solution, we should look at all its subsets. However, the use of a neural network helps to get closer results without checking all subsets of a particular instance and avoids exponential count of such subsets.

\subsection{Variable Discretization}

To achieve a differentiable loss function, a relaxed TSP problem is used, and the optimal solution of such a relaxation does not always correspond to integer solutions \cite{hougardy2014integrality}. To ensure that the solutions provided by minimization of the unsupervised loss function are integer or can be easily converted to integer solutions we use the Gumbel-Softmax technique \cite{Gumbel1,Gumbel2}. We add Gumbel noise of a certain magnitude to logits and then apply softmax to get $x_{ij}$. To understand why adding noise helps to obtain integer solutions, consider how noise affects the  $x_{ij}$. When logits are near zero ($x_{ij}=0.5$) the addition of noise will produce large fluctuations in $x_{ij}$ and consequently large and random loss. But when logits are large ($x_{ij}$ close to integer points) softmax is saturated and the impact of noise becomes negligible. Therefore to minimize the expected loss, the network will be encouraged to produce integer or near-integer solutions. This principle has been described in more detail in other works  \cite{Frey1996ContinuousSB}. 

Since we use greedy search to read out the obtained solution we are not required to fully discretize the solution and noise addition is performed only to improve the solution quality. We experimentally found out that in Euclidean TSP case the best results are obtained without noise but in Asymmetric TSP noise of magnitude 0.1 works well.

\subsection{Implementation}

Implementation of the first three summands of $\mathcal{L}$ is straightforward.
Our approach to efficiently implement the last summand and the parametric connectivity heuristic is based on a known method \cite{2003ImplementingDFJ}. To find the $n$ subsets with the largest $\epsilon$ parameters and therefore the smallest expected cuts, we start with an empty graph $X_0=\left(\{1, ..., n\}, \varnothing \right)$. By adding edges of the proposed solution $X$ to $X_0$ in a descending edge weight order, components of $X_0$ are subsets of $X$ whose $\epsilon$ is at least the value of the last edge added. Hence every time the addition of an edge changes components of $X_0$, the new component corresponds to the subset with the next largest~$\epsilon$. This way we can bypass the explicit calculation of ~$\epsilon$.

For the implementation of this algorithm, only the list of components of $X_0$ is needed. At first, each vertex is in its own component. If the endpoints of the edge with the next largest weight belong to different components, they are merged together by changing the component list; then the cut of the vertex subset of the new component is checked. If the cut sum value in either direction is smaller than one, this new component is insufficiently connected to the rest of the graph and forms a subtour. This is repeated until there are only two components left. 

\begin{algorithm}[h!]
\small
\caption{Parametric connectivity}
\textbf{Input:} Adjacency matrix $X$, where weight of each edge describes if that edge is in the optimal tour; number of vertices $n$ \\

\begin{algorithmic}[1]
\State Initialize empty set of cuts that violate subtour constraint $S$
\State Initialize array of component indexes for each vertex $comp$; $comp \leftarrow [1..n]$
\State Initialize counter $i$; $i \leftarrow 0$
\item[]
\State Add all edges to $E_{decr}$
\State Sort $E_{decr}$ in decreasing order of edge weight
\item[]
\For{$edge$ in $E_{decr}$}
    \item[]
    \Comment{Find components of both $edge$ endpoints}
    \State $c \leftarrow comp[edge\mathrm{.endpoint_1}]$
    \State $c_0 \leftarrow comp[edge\mathrm{.endpoint_2}]$
    \item[]
    \If{$c$ != $c_0$}
        \State $i \leftarrow i+1$ \Comment{Merge components together}
        \For{$v \leftarrow 1$ to $n$} 
            \If{$comp[v]$ is $c_0$}
                \State $comp[v] \leftarrow c$
            \EndIf
        \EndFor
        \item[]
        \State Initialize cut values for each direction $cut_{in}$, $cut_{out}$
        \State $cut_{in}$, $cut_{out} \leftarrow 0$
        \For{$v_1 \leftarrow 1$ to $n$}
            \For{$v_2 \leftarrow 1$ to $n$}
                \item[]
                \Comment{Find values of cuts in both directions}
                \State $c_1 \leftarrow comp[v_1]$
                \State $c_2 \leftarrow comp[v_2]$
                \If{$c_1$ == $c$ and $c_2$ != $c$}
                    \State $cut_{in} \leftarrow cut_{in} + X[v_1, v_2]$
                \EndIf
                \If{$c_2$ == $c$ and $c_1$ != $c$}
                    \State $cut_{out} \leftarrow cut_{out} + X[v_2, v_1]$
                \EndIf
            \EndFor
        \EndFor
        \item[]
        \item[]
        \Comment{Add a subset of vertices (cut) for each violated constraint}
        \For{$sum$ in $[c_{in}, c_{out}]$} 
            \If{$sum < 1$}
                \State Initialize empty set $Q$
                \For{$v \leftarrow 1$ to $n$}
                    \If{$comp[v]$ == $c$}
                        \State $Q \leftarrow Q \cup \{v\}$
                    \EndIf
                \EndFor
                \State $S \leftarrow S \cup \{Q\}$
            \EndIf
        \EndFor
        \If{$i$ = $n-2$}
            \Return $S$
        \EndIf
    \EndIf
\EndFor
\end{algorithmic}
\label{alg:parametric}
\end{algorithm}

To make our method easily usable for both symmetric and asymmetric graphs, all graphs are implemented as directed;  when ordering edge weights and adding edges to $X_0$, a sum of both direction weights is used. However, when observing cuts, each direction is checked separately as this can provide more information.

It has been shown  that parametric connectivity heuristic can be implemented with complexity $O(n^2\alpha(n^2))$ \cite{2003ImplementingDFJ}, $\alpha$ - the inverse of the Ackermann function. However, since the parametric connectivity heuristic is used only when training the model, a more advanced implementation is not necessarily needed. Our implementation of this algorithm is simpler and the complexity of our approach is $O(n^3)$ (see Algorithm \ref{alg:parametric}).

\section{Neural Model}

Our model is based on Joshi et al. \cite{joshi2020learning} and follows the same pipeline: the input is a graph represented by its adjacency matrix $C$ and we train our model to output matrix $X~\in~[0, 1]^{n \times n}$. This matrix shows which edges belong to the optimal tour and its values can be viewed as probabilities. We use a graph neural network that uses both edges and nodes, and each edge and node of the graph is embedded as $d$-dimensional vector. Lastly, we use our differentiable loss function $\mathcal{L}$ to train the network. 

To get the full predicted tours from $X$ (e.g. to use for evaluation), we use greedy search. Greedy search finds the complete tour by starting from a random node and traversing along the heaviest edge which is available until a Hamiltonian cycle is formed.

We also need neural models with supervised learning and reinforcement learning to compare our approach to different learning paradigms. To ensure fairness, we use the same SL and RL models as Joshi et al. \cite{joshi2020learning} used in his work.

\subsection{Graph Neural Network}

Our graph neural network (GNN) consists of several layers and each layer $\ell$ consists of message passing and updating of node and edge embeddings. For initial node embeddings $h_{ij}^{\ell=0}$ and edge embeddings $e_{ij}^{\ell=0}$ we use $d$-dimensional linear projections of node coordinates and normalized edge weights respectively:
\begin{align*}
    & h_{ij}^{\ell=0} = node_{ij} \cdot \mathrm{W_1} + b_1 \numberthis\\
    & norm\_c_{ij} = \frac{ c_{ij}}{\sqrt{\frac{1}{n}\sum_k^n \sum_l^n \left( c_{kl} \right)^2 }} \numberthis\\
    & e_{ij}^{\ell=0} = norm\_c_{ij} \cdot \mathrm{W_2} + b_2 \numberthis
\end{align*}

To update edge and node features message passing is used. Two types of messages are computed from each edge (outgoing from vertex and incoming to vertex) using simple $\mathrm{MLP}$ networks, after that vertices gather and process all messages from their adjacent edges and the vertex itself:
\begin{align*}
    & out\_state_{ij} = \frac{\sum_{k=1}^n  (\mathrm{MLP}_1(e_{ik}^\ell))}{\sqrt{n}} \numberthis\\
    & in\_state_{ij} = \frac{\sum_{k=1}^n (\mathrm{MLP}_2(e_{kj}^\ell))}{\sqrt{n}} \numberthis\\
    & vertex\_state_{ij} = [in\_state, out\_state, h_{ij}^\ell] \numberthis
\end{align*}

Next a new edge embedding candidate for each edge is obtained from the processed messages of adjacent vertices, and the embedding of each edge is updated by combining the old embedding with the new candidate ($tile_{ij}$ is used for ease of implementation):
\begin{align*}
    & tile_{ij} = \begin{array}{c@{\!\!\!}l}
    \left[ \begin{array}[c]{c}
    vertex_{ij} \\
    \vdots \\
    vertex_{ij} \\
    \end{array}  \right]
    & \begin{array}[c]{@{}l@{\,}l}
    \left. \begin{array}{c} \vphantom{0}  \\ \vphantom{\vdots}
    \\ \vphantom{0} \end{array} \right\} & \text{$n$ times}
    \end{array}
    \end{array} \numberthis \\
    & candidate_{ij} = \mathrm{MLP}_3([e_{ij}^\ell, tile_{ij}, tile_{ij}^T]) \numberthis\\
    & e_{ij}^{\ell+1} = e_{ij}^\ell \cdot \sigma(a \cdot \mathrm{A}^\ell) + \mathrm{B}^\ell \cdot candidate \numberthis
\end{align*}

Lastly, the updating of node embeddings $h_{ij}^\ell$ is done by using $\mathrm{MLP}$ network on information available to vertices:
\begin{align*}
    h_{ij}^{\ell+1} = \mathrm{MLP}_4(h_{ij}^\ell) \numberthis
\end{align*}

Each layer contains multilayer perceptrons $\mathrm{MLP}_1$, $\mathrm{MLP}_2$, $\mathrm{MLP}_3$, $\mathrm{MLP}_4$, each with 3 layers (including input and output layer); learnable parameters $\mathrm{W_1}$, $\mathrm{W_2}$, $\mathrm{A}$, $\mathrm{B} \in \mathbb{R}^d$ and $b_1$, $b_2 \in \mathbb{R}$ as well as a scalar value $a$ (we experimentally determined $a = 10$ to work well).

To decode edge embeddings from the last layer of GNN and get probabilities for each edge to belong to optimal tour, we first use two layer MLP to get logits from embeddings; after that we get probability matrix $X$ from logits via softmax over each edge.

If a symmetrical TSP variant is being tackled (e.g. Euclidean TSP), we make logits symmetrical before softmax by taking the mean of logits in each edge direction.

\section{Evaluation}

We carry out several experiments to compare our unsupervised approach to both supervised and reinforcement approaches. 

All datasets are generated randomly. For symmetric graphs, we choose points in a unit square and get adjacency matrices as Euclidean distances between those points. For asymmetric graphs, random adjacency matrices are generated with each edge weight ranging from $0$ to $1$. Correct solution tours (required for evaluation and supervised learning) are computed using \textit{Concorde} solver \cite{Concorde} and \textit{Gurobi} optimizer \cite{Gurobi} for symmetric and asymmetric cases respectively.

We explore all methods on fixed-size graphs of 20 and 50 vertices on Euclidean TSP and on asymmetric TSP. For unsupervised and reinforcement learning training $128000$ examples are randomly generated in each of the 100 epochs; for supervised learning, a larger set of $1280000$ samples and their solutions are generated beforehand. For evaluation, $1280$ samples and their solutions are generated of each type, i.e. TSP and ATSP on respective graph sizes.

To fairly compare between different paradigms, supervised and unsupervised models differ only in the loss function. Comparison with reinforcement learning is not as straightforward considering that the RL model is auto-regressive and builds the solution step by step as opposed to SL and UL models, which are non-autoregressive and produce the solution in one shot. To ensure the comparison is as fair as possible, for RL we use the corresponding encoder described in Joshi et al. \cite{joshi2020learning}. Unfortunately, this means that for asymmetric TSP this encoder has to embed the adjacency matrix into node embeddings. Nonetheless, Table \ref{tab:parameters} contains a summary of the main hyperparameters used for comparing SL, RL, and our UL method. We also follow the experimental setup of Joshi et al. \cite{joshi2020learning}, but some parameters have been adjusted for hardware limitations.

\begin{table}[h!]
\vspace{-5mm}
    \caption{Training parametrs for SL, UL and RL models.}
    \centering
        \begin{tabular}{l|c|c|c}
    \toprule
    Parameter & SL & UL & RL \\
    \midrule
        Epochs & $1$ & $100$ & $100$\\
        Epoch size & $12800000$ & $128000$ & $128000$ \\
        Batch size ($n = 20, 50$) & $128, 32$ & $128, 32$ & $128, 32$ \\
        Encoder layers ($n = 20, 50$) & $16, 8$ & $16, 8$ & $16, 8$ \\
        Number of parameters & $354562$ & $354562$ & $379072$ \\
        Learning rate & $10^{-4}$ & $10^{-4}$ & $10^{-4}$ \\
        Embedding and hidden dimensions & $64$ & $64$ & $64$ \\
    \bottomrule
    \end{tabular}
    \label{tab:parameters}
\end{table}

The output of the model is the probabilities of edges to belong to the correct tour; hence, the greedy search method is used to get a valid tour prediction. Results are compared using optimality gap, i.e. the average percentage ratio between the predicted tour and the correct one. We also look at inference time (1280 samples) and inspect the consistency of validation results throughout training to observe any unstable behaviours. 

Tables \ref{tab:results_tsp} and \ref{tab:results_atsp} shows results of solving Euclidean and asymmetric TSP using different learning paradigms. We evaluate all methods on fixed-size graphs of 20 and 50 vertices.

\begin{table}[h!]
\vspace{-5mm}
    \caption{Optimality gap and inference time  of TSP using SL, RL and UL}
    \centering
        \begin{tabular}{l||cc|cc}
    \toprule
    \multirow{2}{*}{Method} & \multicolumn{2}{c|}{TSP20} & \multicolumn{2}{c}{TSP50} \\
     &  Opt. Gap & Time & Opt. Gap & Time  \\
    \midrule
    SL  & $0.219$ & $2.914$ & $4.870$ & $8.141$  \\
    RL  & $2.752$ & $3.163$ & $7.954$ & $8.881$  \\
    UL  & $1.289$ & $2.852$ & $11.419$ & $8.468$ \\
    \bottomrule
    \end{tabular}
    \label{tab:results_tsp}
\end{table}

\begin{table}[h!]
    \caption{Optimality gap and inference time of ATSP using SL, RL and UL}
    \centering
        \begin{tabular}{l||cc|cc}
    \toprule
    \multirow{2}{*}{Method}  & \multicolumn{2}{c|}{ATSP20} & \multicolumn{2}{c}{ATSP50}  \\
     & Opt. Gap & Time & Opt. Gap & Time \\
    \midrule
    SL  & $17.640$ & $3.225$ & $83.377$ & $9.598$ \\
    RL  & $534.820$ & $3.488$ & $1439.005$ & $9.208$ \\
    UL  & $20.560$ & $3.446$ & $32.699$ & $9.392$\\
    \bottomrule
    \end{tabular}
    \label{tab:results_atsp}
\end{table}

In all of the experiments supervised learning shows superior results to other learning paradigms, which we explain by our experiments being of fixed-sized graphs and supervised learning having all the information of training instances. This also coincides with the current literature \cite{Joshi2019OnLP}.

When comparing unsupervised learning with reinforcement learning, we can observe that results on Euclidean instances are ambiguous, as UL performs better on 20 vertices, but RL surpasses UL when looking at instances with 50 vertices. However, results on asymmetric TSP are much more certain, where RL behaves very poorly. This suggests that the encoder used is not powerful enough to efficiently embed the adjacency matrix into node embeddings.

When comparing results between TSP and ATSP, we can see that performance for asymmetric instances is noticeably worse as it is a generally harder problem. However, UL achieves similar results to SL for asymmetric TSP, which may indicate the adaptiveness of the unsupervised approach.

Inference time results between learning paradigms are very similar and no noteworthy differences can be seen.

\begin{figure}[h!]
\centering 
\subfloat[TSP20\label{fig:tsp20}]{\includegraphics[width=0.5\textwidth]{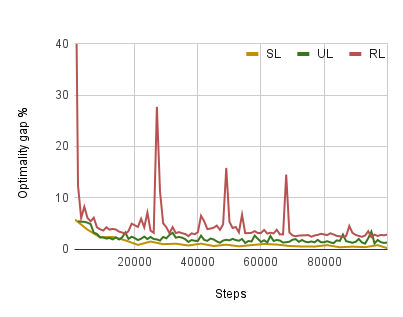}}\hfill
\subfloat[TSP50\label{fig:tsp50}] {\includegraphics[width=0.5\textwidth]{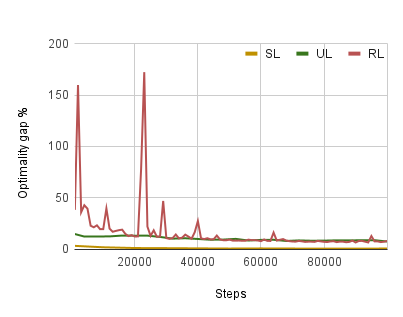}}\hfill

\subfloat[ATSP20\label{fig:atsp20}]{\includegraphics[width=0.5\textwidth]{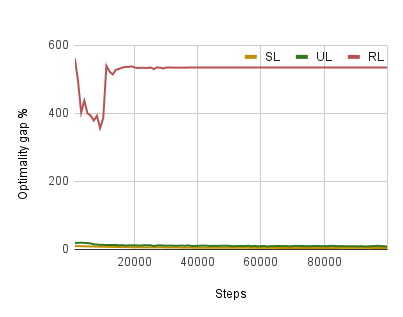}}\hfill
\subfloat[ATSP50\label{fig:atsp50}] {\includegraphics[width=0.5\textwidth]{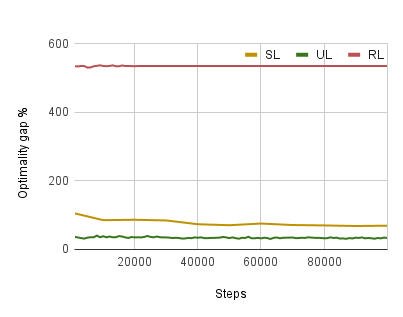}}\hfill
\caption{Comparison of optimality gap throughout training when using RL, SL and UL} \label{fig:training}
\end{figure}

Figure \ref{fig:training} shows training behaviors of each of the learning paradigms in all experiments carried out (validation done with greedy search). It can be seen in Euclidean experiments (Figure \ref{fig:tsp20}, Figure \ref{fig:tsp50}) that reinforcement learning has several big fluctuations in the training process; we do not experience this with other learning paradigms. This type of unstable behaviour is a relatively more common behaviour for RL in general and leads to large amount of steps to train the neural network properly. The asymmetric training graphs (Figure \ref{fig:atsp20}, Figure \ref{fig:atsp50}) show small or no improvement in RL training over time, indicating that the encoder used is not suitable for asymmetric TSP.

To better see how unsupervised loss work with neural networks, we carried out an experiment to compare straightforward minimization of the loss function and its usage in our model. The minimization of the function was done using Adam optimizer ($\textit{learning rate} = 0.01$) and we let it run on each instance of the evaluation datasets for $15000$ steps which were empirically determined to be enough for most edges to be almost discrete. To get proper tours from the output, we use greedy search.

Results in this experiments for Euclidean TSP and asymmetric TSP can be seen in in Table \ref{tab:lossTSP} and Table \ref{tab:lossATSP}. We tracked both the optimality gap for each method as well as inference time for 1280 instances. It should be noted that time spent on loss minimization directly depends on optimization steps and could be reduced by possibly sacrificing the quality of the solution. For better comprehension of the experiment, we added average results of a random tour and also of a tour found with greedy search on the adjacency matrix.

\begin{table}[b!]
\vspace{-5mm}
    \caption{Comparison of minimization of loss function and loss function used in neural network for TSP}
    \centering
        \begin{tabular}{l||cc|cc}
    \toprule
    \multirow{2}{*}{Method}  & \multicolumn{2}{c|}{TSP20} & \multicolumn{2}{c}{TSP50} \\
     & Opt. Gap & Time & Opt. Gap & Time \\
     \midrule
    Random &  $186.959$ & $0.007$ & $371.908$ &  $0.011$  \\
    Greedy search &  $17.620$ & $0.169$ & $22.801$ & $1.220$   \\
    Loss function &  $6.736$ & $25181.865$ & $16.560$ & $54313.828$   \\
    Neural network &  $1.289$ & $2.852$ & $11.419$ & $8.468$ \\
    \bottomrule
    \end{tabular}
    \label{tab:lossTSP}
\end{table}

\begin{table}[h!]
    \caption{Comparison of minimization of loss function and loss function used in neural network for ATSP}
    \centering
        \begin{tabular}{l||cc|cc}
    \toprule
    \multirow{2}{*}{Method} &  \multicolumn{2}{c|}{ATSP20} & \multicolumn{2}{c}{ATSP50}  \\
     & Opt. Gap & Time & Opt. Gap & Time \\
    \midrule
    Random &  $556.233$ & $0.007$ & $1492.072$ &  $0.0011$ \\
    Greedy search &   $91.194$ & $0.169$ & $145.298$ & $1.220$ \\
    Loss function &   $26.413$ & $24448.681$ & $47.762$  & $54002.372$ \\
    Neural network &  $20.560$ & $3.446$ & $32.699$  & $9.392$ \\
    \bottomrule
    \end{tabular}
    \label{tab:lossATSP}
\end{table}

As expected, we can see that the minimization of the loss function returns better results than just greedy search. When the loss function is used together with a neural network, the results are even better. This can be explained by the fact that the loss function has many local minimums in which the optimizer can get trapped, but a neural network helps to overcome this. If we look at the inference times, we can see that individual optimization is very slow and is not practical for widespread use.

\section{Conclusions}

We introduce a novel unsupervised learning approach for solving the TSP problem with neural networks. The basis of our unsupervised method is a new differentiable loss function that works on both Euclidean and asymmetric TSP. Unsupervised learning has the advantage over supervised learning of not needing large correctly labeled datasets. Our method performs similarly to reinforcement learning for Euclidean graphs with 20 and 50 vertices and outperforms reinforcement learning when looking at the stability of training or asymmetric graphs. 

The loss function is constructed in a way to be easily modified with extra constraints. The addition of constraints can be done by expressing the constraint as a differentiable polynomial and adding it to the loss function. Considering that routing problems are very widespread and often have unique limitations, the addition of constraints is very relevant and may be very useful. This work does not explore this possibility further but in the future, we want to examine our work on TSP variants with additional constraints.

\bibliography{main.bib}
\end{document}